# An Analysis System for DNA Gel Electrophoresis Images Based on Automatic Thresholding and Enhancement

Naima Kaabouch[1], *Member, IEEE*, Richard R. Schultz[1], *Member, IEEE*, Barry Milavetz[2]
[1]Department of Electrical Engineering
[2]Department of Biochemistry and Molecular Biology
School of Medicine and Health Sciences
University of North Dakota, Grand Forks, ND 58202-7165

*Abstract*-Gel electrophoresis, a widely used technique to separate DNA according to their size and weight, generates images that can be analyzed automatically. Manual or semiautomatic image processing presents a bottleneck for further development and leads to reproducibility issues. In this paper, we present a fully automated system with high accuracy for analyzing DNA and proteins. The proposed algorithm consists of four main steps: automatic thresholding, shifting, filtering, and data processing. Automatic thresholding, used to equalize the gray values of the gel electrophoresis image background, is one of the novel operations in this algorithm. Enhancement is also used to improve poor quality images that have faint DNA bands. Experimental results show that the proposed technique eliminates defects due to noise for average quality gel electrophoresis images, while it also improves the quality of poor images.

## 1. Introduction

In the past decade, in addition to several commercial software packages for filtering and identifying the gel bands such as Scanalytics™, GelcomparII™, Gel-Pro Analyzer™, and TotalLab™, many methods [1-6] have been proposed for filtering, segmenting, and detecting gel bands, as well as rectifying their geometries. These methods have two major disadvantages. Some use semi-automatic techniques and require the manual setting of some sensitivity threshold related to the grayscale intensity of the bands. End-users must often adjust these parameters for individual bands. Other systems automatically filter and smooth grayscale intensities, often causing some true bands to disappear into the background while some false bands remain in the image and have to be deleted manually by the user. The fact that the user has to adjust or delete these bands and manually compute some property such as blob area by using rectangular markers can lead to the problem of reproducibility, in addition to being rather time consuming for the bench scientist. Because the existing approaches are not able to achieve automatic computation and high accuracy, in a pervious paper we proposed an alternative method for gel electrophoresis image analysis by designing an automated system that is free of user intervention [7]. In this paper, we improved this automated system that uses new algorithm to achieve high accuracy and reproducibility for DNA data analysis. The method automatically computes key parameters, such as the gel band size and center of mass. The system also compares the size of extracted bands to a selectable reference band.

## 2. Quality of gel images

Many gel electrophoresis images frequently contain anomalies such as salt and pepper noise and large smears on a strong non-uniform background. Although the DNA bands are all visible, some are faint, and their gray levels may be close to the background level. Some of them have a long-tailed shape, which makes difficult to distinguish the limits of the DNA bands.

Examples of such gel images are shown in Figs. 1 and 2, with Fig. 1 containing simplex DNA bands, and Fig. 2 representing a duplex DNA pattern. Both of these gel images have non-uniform backgrounds, noisy stains, and long-tailed smeared DNA bands. In addition to these problems, some of the bands are faint or are too close to each other in the case of duplex DNA.

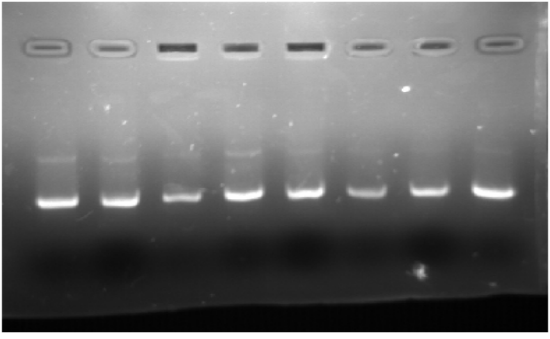

Fig. 1. Example of simplex DNA gel electrophoresis image.

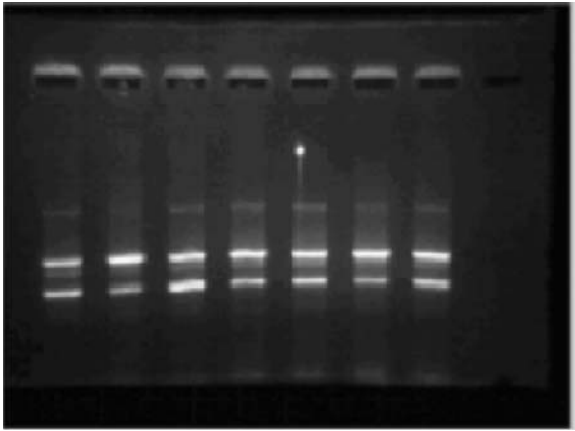

Fig. 2. Example of duplex DNA gel electrophoresis image.

3. METHODOLOGY

The algorithm involves mainly four steps: automatic thresholding, shifting and filtering, detecting and annotating gel bands, and data processing. If some of DNA bands are faint and the system fails to detect all of the bands, an enhancement is then performed.

A graphical user interface was created, allowing the user to open and display any 2-D gel electrophoresis image from a directory and automatically run the software. A final filtered image is displayed, and the user has the option to choose a reference spot and a second DNA spot, as well as to compute and display their area ratios. Other parameters can also be computed and displayed. If the end-user estimates that not all of the bands were detected in the final image, it is possible to perform a pre-enhancement and run the software on the newly enhanced image.

*3.1. Automatic thresholding*

Various factors, such as nonstationary and correlated noise, ambient illumination, busyness of gray levels within the object and its background, inadequate contrast, and experimental errors complicate the thresholding operation. Automatic thresholding has been addressed in a number of papers [8-19]. These techniques proposed were classified by Sezgin et al. [8] in six groups according to the information they are exploiting. These categories are:

1. Histogram shape-based methods, where, for example, the peaks, valleys and curvatures of the smoothed histogram are analyzed [9-10],
2. Clustering-based methods, where the gray-level samples are clustered in two parts as background and foreground object, or alternately are modeled as a mixture of two Gaussians [11-12],
3. Entropy-based methods use the entropy of the foreground and background regions, and the cross-entropy between the original and binarized image [13-14],
4. Object attribute-based methods search a measure of similarity between the gray-level and the binarized images, such as fuzzy shape similarity, and edge coincidence [15-16]
5. Spatial methods use higher-order probability distribution and/or correlation between pixels [17-18],
6. Local methods adapt the threshold value on each pixel to the local image characteristics [19-20].

Due to the diversity and quality of images, the existing thresholding techniques still require considerable human intervention and pre-assumptions to determine appropriate threshold values. In addition to this problem, most of these techniques are not suitable for DNA images because they alter the size of the bands after thresholding.

The purpose of our automatic thresholding algorithm is to equalize the grayscale levels of the gel electrophoresis image background without affecting the size of the bands according to the following equation:

$$I_{th}(x,y) = \begin{cases} I(x,y) & \text{if } I(x,y) > Th\_level \\ Th\_level & \text{otherwise} \end{cases} \quad (1)$$

In this case, $I$ is the matrix representing the input gel image, $I_{th}$ is the matrix representing the output image after thresholding, and $Th\_level$ is the threshold level. The threshold level is a function of the maximum and the minimum of the gel image intensities. For each gel image, the value of the threshold level is computed as

$$Th\_level = \alpha_i [\max(I) - \min(I)] + \min(I), \quad (2)$$

Where $I$ is the matrix representing the gel image to be thresholded, and $\alpha_i$ is a weighting value less than 1. In order to find the best value of the parameter $\alpha_i$ that is suitable for most gel images captured by a single camera, we used the standard deviation profile. The profile of the standard deviation is tresholded by using a threshold level that is between the minimum of the peaks and the minimum value of the standard deviation. This level is then used to compute a parameter $\alpha_s$ given by

$$\alpha_s = (Th\_level_{std} - \sigma_{min})/(\sigma_{max} - \sigma_{min}) \qquad (3)$$

Here, $Th\_level_{std}$ is the threshold level of the standard deviation, $\sigma_{max}$ is its maximum value, and $\sigma_{min}$ is its minimum value. Assuming that $\alpha_i$ from the equation (2) is equal to $\alpha_s$ from equation (3), the value of this last parameter is then used to shreshold the grayscale intensity of each pixel according to the equation (2).

A typical profile is given by Fig. 3 corresponding to the original image of Fig. 2, where the threshold level of the standard deviation is represented by the broken line. Fig. 4 represents the output image after the thresholding operation of the image 2. As can be observed from this image the background is more uniform than before thresholding.

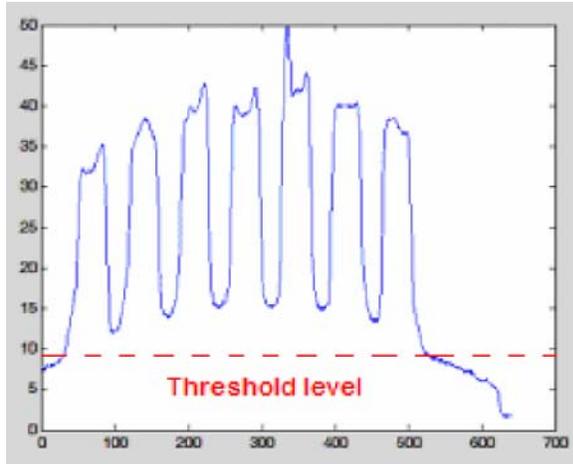

Fig. 3. Profile from the original image in Fig. 2.

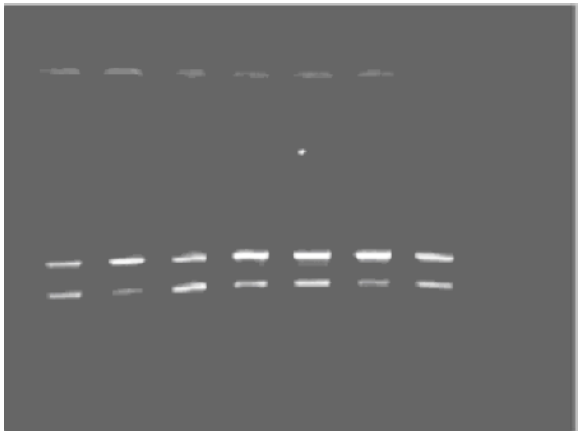

Fig. 4. Output image of the duplex DNA image of Fig. 2 after thresholding.

### 3.2. Shifting and filtering

The objective of this part of the algorithm is to shift the minimum level of the profile to zero and to remove as much noise as possible from the gel electrophoresis image. The shifting operation performs a subtraction of the automatic thresholding level from the intensity of every pixel:

$$I_{sh}(x,y) = I_{th}(x,y) - Th\_level \qquad (4)$$

Here, $I_{sh}$ is the matrix representing the output image after shifting, $I_{th}$ is the matrix representing the output image after thresholding, and $Th\_level$ is the threshold level.

Fig. 5 shows a typical profile after shifting. As observed, the gray level of each pixel is decreased by the threshold level. The background is darker than before shifting, but the amplitude of the peaks (i.e. maximum-minimum) is not affected by the shifting operation. However, because of the noise, the minimum of these peaks is not linear. After this shifting operation, to speed up the computation time and because the injection wells is not needed for this part of the algorithm, we used a nonlinear filter to isolate the area of interest. Since the background is darker, to remove any small bright noise, we used a top hat filter with a morphological structuring element of 10 pixels. However, this filter was not enough to remove all remaining noise, even with repeated application. Applying this filter many times actually altered the size of the DNA bands. As a result, we used the top-hat filter followed by a nonlinear filter, a 5x5 median, to remove any remaining salt-and-pepper noise. Fig. 6 shows an example of the resulting profile after the filtering operation. The gray level of the minimum is near zero and hence the background is darker as shown in Figs. 7 and 8. The DNA spots are also darker, but their sizes are slightly, if at all, affected by these operations.

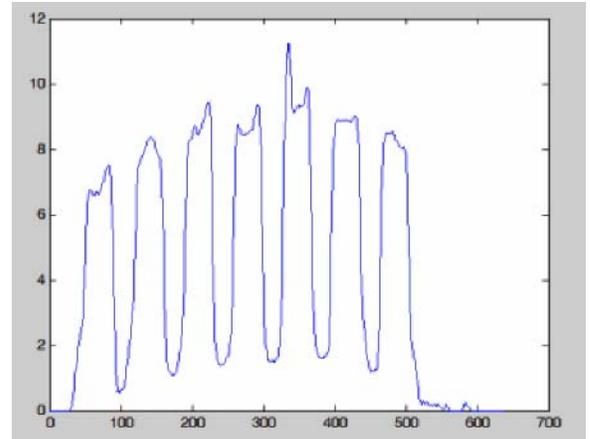

Fig. 5. Profile from the output after thresholding and shifting.

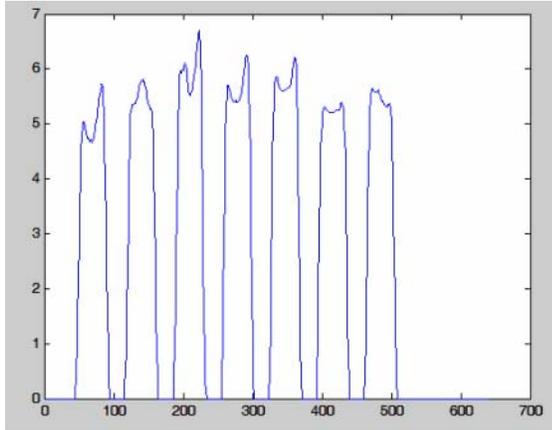

Fig. 6. Profile from the output after thresholding and shifting.

### 3.3. Enhancement

Although it is not applied on all gel electrophoresis images, enhancement is also one of the important processes in this algorithm. It improves the quality of poor gel images and increases the accuracy of the system. Its main objective is to highlight faint bands that can be washed out by the thresholding process because of their low gray levels. Several enhancement techniques have been studied in the past [21-26]. We have considered many of these methods including histogram equalization. However, with histogram equalization, the gray level is distributed so that the resulting histogram has a uniform distribution. For the gel images, the goal is not to create a uniform gray level but to enhance the difference between the DNA bands and the background. Background subtraction was also considered, but the enhanced gel images by this technique showed insignificant improvement. Other nonlinear enhancement techniques were also considered and discarded, mainly because when using these techniques, the DNA bands and the small noisy spots are enhanced, increasing the quantity and the size of noisy spots in the gel images. To solve this problem, an indirect method that consists of a combination of top-hat and bottom-hat filtering has been used. The top-hat filter subtracts an opened image from the original image, while the bottom-hat transformation is defined as the closing of the image minus the image. When combined, these two transformations enhance the difference between dark and bright bands. The operation is given as

$$I_{enh}(x,y) = [\, I_{top}(x,y) + I(x,y) \,] - I_{bot}(x,y), \quad (5)$$

where $I_{enh}$ is the matrix representing the output image after enhancement, $I_{top}$ is the matrix representing the output image after top-hat filtering, $I$ is the matrix representing the original image, and $I_{bot}$ is the matrix representing the output image after shifting.

### 3.4. Data processing

Since the technique of thresholding and filtering remove all the noise, object detection is the appropriate technique to detect all the bands. Quantitative information, such as the amount of substance in each band and the molecular weight of each band, is computed by calculating the area of the band, and by considering the position relative to a predefined reference band, respectively. The ratio size between two selected bands, a reference and another spot is also computed according to the following equation:

$$Ratio\_size\,(n) = area(n) / [area(ref) + area(n)] \quad (6)$$

Here, $n$ is the DNA spot number, and Ref is the DNA reference number. The size of each band is computed by adding all the pixels existing within this band.
The computed data, along with other information, are saved in a file for possible further processing.

### 4. RESULTS

To test the performance and accuracy of the system, we accessed a database of 60 gel electrophoresis images. A total of 29 of these images contained simplex DNA bands, and 17 images contained duplex DNA bands, all of which were assumed to be exploitable. A total of 14 gel images were of very poor quality (i.e., smeared), and it was assumed that these images could not be analyzed in either an automatic or manual fashion. Fig. 7 shows the output image of Fig. 1, after automatic thresholding, shifting, and filtering. Fig. 8 shows the final processed image of the Fig. 2. As shown completely, all of the bands are fully detected and the noise is removed.

The evaluation of the system was based on two criteria, namely full detection and recognition with respect to the DNA size. Table 1 gives a summary of the classified images and the percentage of accuracy for the automated system. As observed in this table, the accuracy of this method is higher than the accuracy of any existing off the shelf system that usually does not exceed 90% for only good quality images. Even for very poor quality data, the system detected all of the DNA bands in 65% of images. The problem of the other six images can be resolved by decreasing the level of thresholding that is calculated automatically.

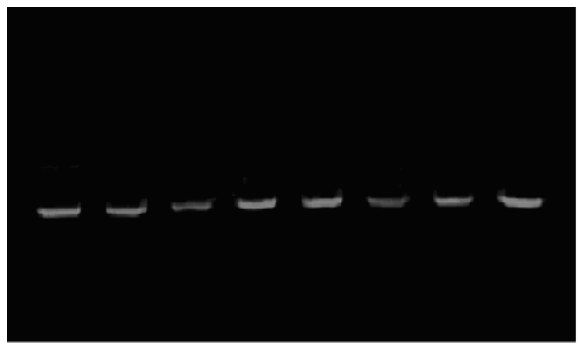

Fig. 7. Output image of the simplex DNA image in Fig. 1.

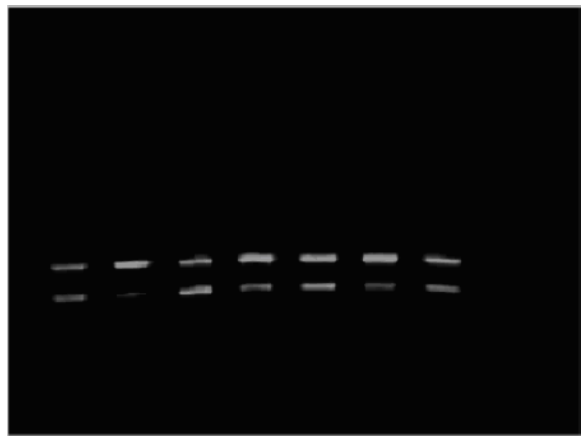

Fig. 8. Output image of the duplex DNA image in Fig. 2.

TABLE 1
Accuracy of the system for the classified gel images.

| Quality | Number of Images | Enhancement | Accuracy |
|---|---|---|---|
| Good | 21 | Not needed | Detection 100% |
| Average | 22 | 8 needed enhancement | Detection 100% |
| Poor | 17 | 13 needed enhancement | 65% Detected |

5. CONCLUSION

We have developed a new system for analyzing DNA gel electrophoresis bands in genetic studies. The proposed algorithm is fully automatic and free of user interaction. This automated system isolates and annotates the DNA bands, as well as computes some parameters related to these bands such as center of mass, size, and ratio-size. This software tool will allow researchers to speed up their studies and achieve more reproducible results.

ACKNOWLEDGMENT

This work was supported by the ND EPSCoR project through National Science Foundation grant # UND0012168.